\documentclass{article}

\usepackage[final]{neurips_wrl2021}

\usepackage[utf8]{inputenc} 
\usepackage[T1]{fontenc}    
\usepackage{hyperref}       
\usepackage{url}            
\usepackage{booktabs}       
\usepackage{amsfonts}       
\usepackage{nicefrac}       
\usepackage{microtype}      
\usepackage{microtype}
\usepackage{graphicx}
\usepackage{subcaption}
\usepackage{booktabs}
\usepackage{algorithm,algorithmic}
\usepackage{amsmath}
\usepackage{float}

\title{Guiding Evolutionary Strategies by Differentiable Robot Simulators}

\author{%
  Vladislav Kurenkov\thanks{Primary author; Personal web-page:  \href{https://vkurenkov.me/}{vkurenkov.me}} \\
  Tinkoff\\
  \texttt{v.kurenkov@tinkoff.ru} \\
   \And
   Bulat Maksudov \\
   Independent Researcher \\
   \texttt{b.maksudov@lua.tatar} \\
}

\begin{document}

\maketitle

\begin{abstract}
In recent years, Evolutionary Strategies were actively explored in robotic tasks for policy search as they provide a simpler alternative to reinforcement learning algorithms. However, this class of algorithms is often claimed to be extremely sample-inefficient. On the other hand, there is a growing interest in Differentiable Robot Simulators (DRS) as they potentially can find successful policies with only a handful of trajectories. But the resulting gradient is not always useful for the first-order optimization. In this work, we demonstrate how DRS gradient can be used in conjunction with Evolutionary Strategies. Preliminary results suggest that this combination can reduce sample complexity of Evolutionary Strategies by 3x-5x times in both simulation and the real world.
\end{abstract}

\section{Introduction}

Evolutionary Strategies is a class of zeroth-order black-box optimization algorithms that were successfully applied in various simulated robotics tasks \citep{maniaSimpleRandomSearch2018, salimansEvolutionStrategiesScalable2017}. They are known as easy to parallelize, and a small number of hyperparameters makes them easy to tune. However, these methods were observed to exhibit a higher sample complexity (in comparison to Reinforcement Learning algorithms) and a strong dependence on an initial network initialization \citep{chrabaszczBackBasicsBenchmarking2018}, making their adoption for training robots directly in reality troublesome.

Recently, there was a rapid growth of work on Differentiable Robot Simulators (DRS) \citep{degraveDifferentiablePhysicsEngine2019, huDiffTaichiDifferentiableProgramming2020, geilingerADDAnalyticallyDifferentiable2020, qiaoScalableDifferentiablePhysics2020, sanchez-gonzalezLearningSimulateComplex2020, deavilabelbute-peresEndtoEndDifferentiablePhysics2018}. The promise is to drastically reduce the number of samples needed for finding successful policies in comparison to reinforcement learning or black-box policy search methods. However, implementing DRS is non-trivial and requires specific approaches (especially for collision handling \citep{geilingerADDAnalyticallyDifferentiable2020, qiaoScalableDifferentiablePhysics2020}) to make the resulting gradient useful for first-order optimization. Moreover, first-order optimization methods that rely on DRS's gradient may get stuck in local optima and some specific treatment could be needed. 

Lately, multiple attempts were made to merge Evolutionary Strategies (ES) and Reinforcement Learning (RL) algorithms \citep{contiImprovingExplorationEvolution2018, khadkaEvolutionGuidedPolicyGradient2018, pourchotCEMRLCombiningEvolutionary2019}. Following this line of work, instead of combining ES and RL, we aim to combine the ends of the expected sample complexity spectrum, namely, DRS and Evolutionary Strategies. Specifically, in cases where the former is not useful for optimization with first-order methods.  We propose to use recently introduced Guided Evolutionary Strategies (Guided-ES) \citep{maheswaranathanGuidedEvolutionaryStrategies2019} and treat DRS's gradient as a surrogate. Using this combination, we demonstrate that it is possible to reduce sample complexity of Evolutionary Strategies by 3x-5x times when training directly on a real robot (Section \ref{sec:exp:accelerate}). Furthermore, we show in simulation that even misleading gradients from a DRS can be utilized to speed up the convergence of Evolutionary Strategies (Section \ref{sec:exp:misleading}).


    

\section{Evolutionary Strategies with Differentiable Robot Simulators}
\label{sec:es_drs}

To unite DRS and Evolutionary Strategies, we propose to use an algorithm introduced in \cite{maheswaranathanGuidedEvolutionaryStrategies2019} -- Guided Evolutionary Strategies. This method can make use of any surrogate gradients that are correlated with the true gradient to accelerate the convergence of evolutionary strategies. The surrogate gradients can be corrupted or biased in any way, the only requirement for them is to preserve a positive correlation with the true gradient. We presume that this is the case for the gradients computed with DRS and therefore propose to use it as a surrogate in Guided-ES.

The proposed approach is outlined in Algorithm \ref{alg:ges-drs-mass-spring}. The key idea is to compute surrogate gradient $\nabla f_{drs}$ using a differentiable robot simulator. Then a history of $k$ previous surrogates can be formed into a matrix of size $n \times k$ (where $n$ is a size of search space) and an orthonormal basis $U$ extracted. This basis is next used to define a covariance matrix $\Sigma$ utilized for generating the perturbations. The hyperparameter $\alpha$ controls how biased should the perturbations be in the direction of the surrogate gradient. 

\begin{algorithm*}[t]
 \caption{Evolutionary Strategies Guided by DRS}
 \label{alg:ges-drs-mass-spring}
 \begin{algorithmic}[1]
 \renewcommand{\algorithmicrequire}{\textbf{Input:}}
 \renewcommand{\algorithmicensure}{\textbf{Output:}}
 \REQUIRE {Initial solution $\theta_{0}$; Optimizer $opt$; Cost function $f(\theta)$; DRS Gradient $\nabla f_{drs}$}
 \ENSURE  Final solution $\theta_{T}$
 \FOR {$t = 0$ to $T$}
  \STATE \textit{// DRS Part}
  \STATE Get DRS gradient $\nabla f_{drs}(\theta_{t})$
  \STATE
  \STATE \textit{// Guided-ES Part}
  \STATE Update low-dimensional guiding subspace $U$ with the DRS gradient
  \STATE Define search covariance $\Sigma = \frac{\alpha}{n}I + \frac{1-\alpha}{k}UU^{T}$
  \FOR {$i = 1$ to $P$}
   \STATE Sample perturbation $\epsilon_{i} \sim \mathcal{N}(0, \sigma^{2}\Sigma)$
   \STATE Compute antithetic pair of losses $f(\theta_{t} + \epsilon_{i})$ and $f(\theta_{t} - \epsilon_{i})$
  \ENDFOR
 \STATE Compute Guided ES gradient estimate $g = \frac{\beta}{2\sigma^{2}P} \sum_{i=1}^{P}\epsilon_{i}[f(\theta_{t} + \epsilon_{i}) - f(\theta_{t} - \epsilon_{i})]$
 \STATE Update parameters using given optimizer $\theta_{t+1} = opt.step(\theta_{t}, g)$
 \ENDFOR
 \RETURN $\theta_{T}$ 
 \end{algorithmic} 
\end{algorithm*}

We note that there is no need to rely on the proposed zeroth-order optimization if one has access to the exact gradient of the objective function. However, there are cases where the gradient obtained with DRS is not effective when used with first-order optimization methods to optimize the objective. In the following section, we will demonstrate two of them and show how one can benefit from the proposed approach to improve the sample efficiency of evolutionary strategies.

\section{Experiments}
\subsection{Accelerated Learning on Real Robot}
\label{sec:exp:accelerate}

Evolutionary strategies and their variants were observed to be on par with reinforcement learning algorithms \citet{salimansEvolutionStrategiesScalable2017}, but require a larger amount of episodes for training in some cases. Still, algorithms of this class were successfully applied to train robots directly in the real world at the expense of higher experimental time \cite{floreanoEvolutionaryRoboticsBiology2000}. Here, we expect that incorporating information from DRS into the training process should accelerate the convergence of evolutionary strategies.

We use the approach described in Section \ref{sec:es_drs}, where a surrogate gradient is taken as a difference between current parameters and parameters obtained after a fixed amount of optimization steps in simulation with DRS. The ascent direction obtained with DRS does not exactly match the ascent direction of the objective function $f_{real}(\theta)$ that depends on the real world, but we expect it to be at least positively correlated. A detailed algorithm can be found in the appendix\footnote{Source code: \href{https://github.com/vkurenkov/guided-es-by-differentiable-simulators}{https://github.com/vkurenkov/guided-es-by-differentiable-simulators}}.

\begin{figure}[!htbp]
\begin{center}
\centerline{\includegraphics[width=\columnwidth]{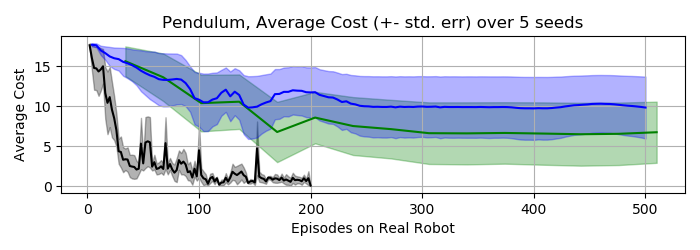}}
\centerline{\includegraphics[width=\columnwidth]{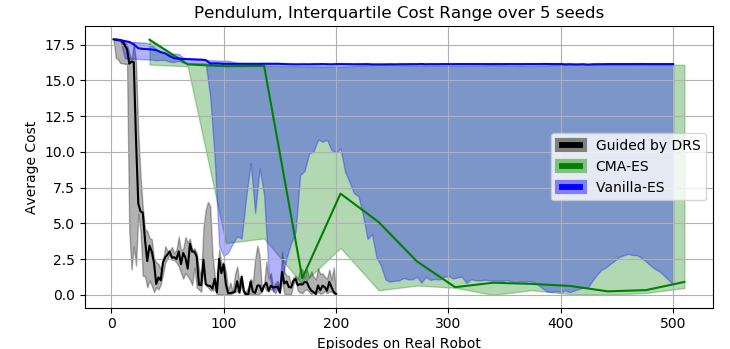}}
\caption{Guided-ES with DRS achieves considerably lower sample complexity for training on real robot in comparison to vanilla evolutionary strategies (Vanilla-ES) and Covariance Matrix Adaptation Evolution Strategy (CMA-ES). Moreover, the method is more robust across random seeds as indicated on the interquartile cost range graph.}
\label{fig:pendulum:comparison}
\end{center}
\end{figure}

Due to limited access to the experimental platform with physical robots, we only consider one problem -- a swinging pendulum \cite{chungNonlinearControlSwinging1995}, where the goal is to start swinging constantly at 180 degrees. The cost function is defined as a sum of differences between current energy and target energy at each timestep. The state space is represented by the last four measurements of position and velocity.

We compare Guided-ES with DRS against Vanilla Evolutionary Strategies (Vanilla-ES) \cite{salimansEvolutionStrategiesScalable2017} and also against Covariance Matrix Adaptation Evolution Strategy (CMA-ES) \cite{hansenReducingTimeComplexity2003} to involve a method that adapts the covariance matrix during the training process. In Figure \ref{fig:pendulum:comparison}, we observe that the proposed method converges faster than both Vanilla-ES and CMA-ES, moreover, the convergence is robust to random seeds in opposition to other considered algorithms. We notice that Vanilla-ES and CMA-ES are highly dependent on initial network initialization (to the extent of non-convergence under our experimental budget), which is not the case for Guided-ES with DRS. 

We find the results of this experiment encouraging as it gives a piece of evidence in favor of leveraging DRS (and a closed-loop transfer from simulation in general) for data-efficient evolutionary strategies. But more sophisticated robots and problems should be probed further.
    
\subsection{When DRS Gradients Are Misleading}
\label{sec:exp:misleading}

    

To demonstrate that the convergence of Guided-ES with DRS is possible even when the gradients are not useful for first-order optimization methods, we rely on a Mass-Spring simulator depicted in Figure \ref{fig:mass_spring}. \cite{huDiffTaichiDifferentiableProgramming2020} observed that a naive implementation of this simulator results in a gradient that can not be used with a stochastic gradient descent -- optimization process does not converge to a satisfactory solution. And a specific approach to collision handling (time-of-impact fix) is necessary. In this setup, we want to probe how well Guided-ES performs if provided with gradients computed using a naive version of the simulator, therefore we remove the fix proposed by \cite{huDiffTaichiDifferentiableProgramming2020}.

\begin{figure}[ht]
\begin{center}
\centerline{\includegraphics[width=\textwidth]{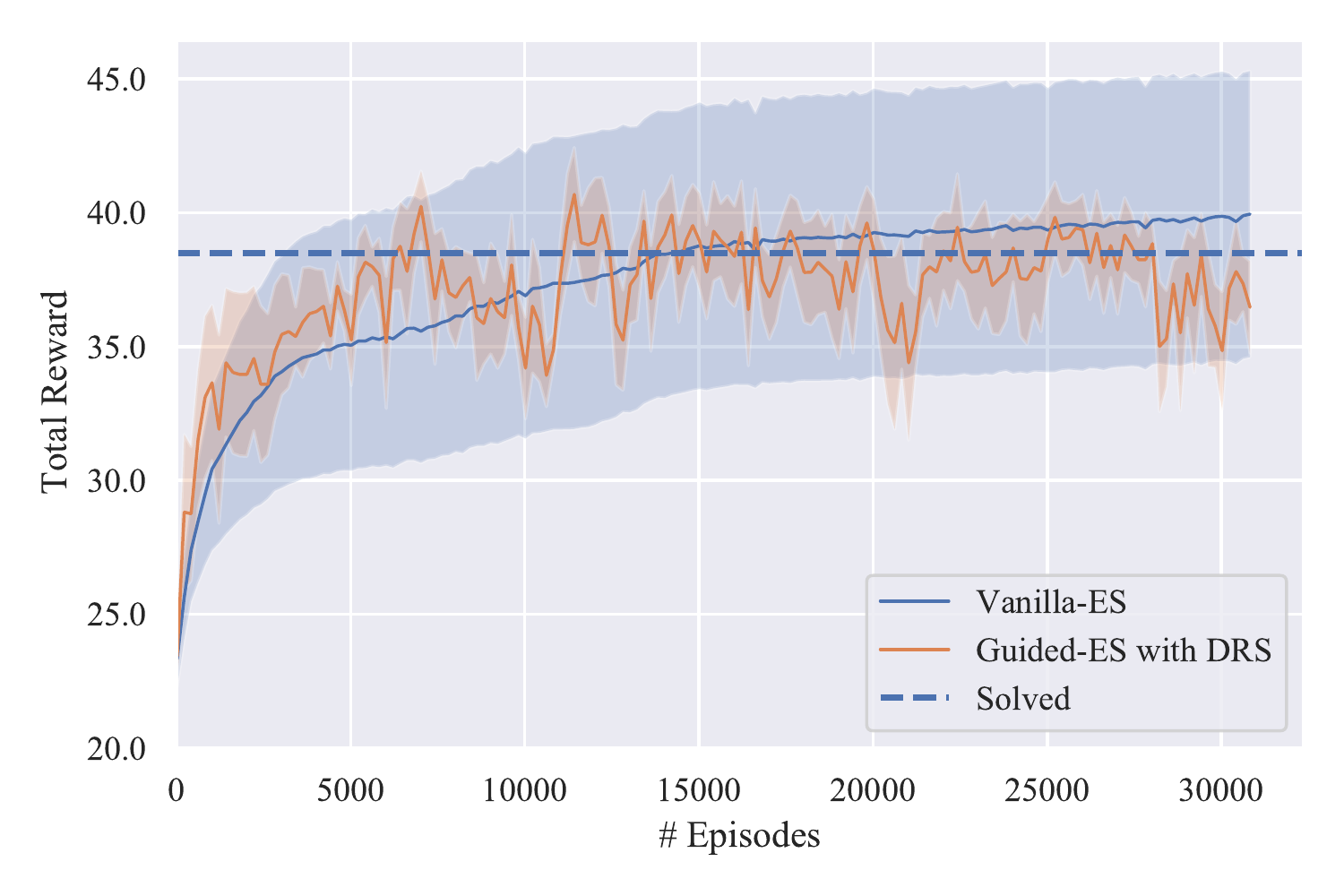}}
\caption{Guided-ES with DRS achieves lower sample complexity in comparison to vanilla evolutionary strategies (Vanilla-ES) with lower in-sample variation. \textit{[Computed across 5 seeds for each algorithm, shaded area represents 95\%-confidence interval.]}}
\label{fig:low_sample}
\end{center}
\end{figure}

We observe in Figure \ref{fig:low_sample} that the proposed approach is able to find a satisfactory solution even when guided by the ineffective gradient. Moreover, the convergence on average is faster than with simple evolutionary strategies. However, we observe that variation between training iterations is quite high, but we believe it can be overcome with proper learning rate scheduling. On the other side, in-sample variation -- within a training iteration but over multiple seeds, is low suggesting that a satisfactory result is found fast independent of a network initialization (which is not the case for Vanilla-ES). The results of this experiment suggest that if one has access to a differentiable robot simulator with inaccurate backward propagation, they should not abandon such a simulator. As there is a possibility it can still be used to reduce sample complexity of evolutionary strategies, which are easier to tune than reinforcement learning algorithms.

\section{Conclusion and Future Work}

In this work, we proposed a natural way to combine Differentiable Robot Simulators and Evolutionary Strategies, and demonstrated two cases where such a combination could be beneficial to reduce sample complexity of the latter. We find these results encouraging, especially if one is interested to train robots directly in real life. However, future work should include more control robotic problems to probe the viability of the proposed method in more intricated setups that involve real and simulated robots.

\section*{Acknowledgments}

The authors would like to thank Oleg Balakhnov and Sergei Savin for helpful discussions and help with the robot hardware.





\bibliography{example}

\section{Appendix}

\subsection{Hyperparameters}
For experiments on the real robot, we did an extensive hyperparameter search for each algorithm in simulation and used the best set of them for training on the real robot. For experiments in Section \ref{sec:exp:misleading} we also did an extensive hyperparameter search and reported the best curves for both of the algorithms. 

We tried several optimization algorithms (SGD, Adam, RMSProp, AdaGrad, Fromage) and found the Fromage \cite{bernsteinDistanceTwoNeural2021} to perform most stable when DRS gradients are involved, we believe this is due to its inherent property of normalizing the gradients, which we observed to be a necessary additional step for all other optimization algorithms to converge with DRS gradient.

    		

\subsection{Pseudocode for Accelerated Learning on Real Robot}
\begin{algorithm*}[h]
 \caption{Evolutionary Strategies Guided by DRS for Training on Real Robot}
 \label{alg:ges-drs-real}
 \begin{algorithmic}[1]
 \renewcommand{\algorithmicrequire}{\textbf{Input:}}
 \renewcommand{\algorithmicensure}{\textbf{Output:}}
 \REQUIRE {Initial solution $\theta_{(0)}$; Optimizers $opt_{real}, opt_{sim}$; Cost functions $f_{real}(\theta)$, $f_{drs}(\theta)$; DRS Gradient $\nabla f_{drs}$}
 \ENSURE  Final solution $x_{T_{real}}$
 \FOR {$t = 0$ to $T_{real}$}
  \STATE \textit{// DRS Part}
  \STATE $\theta_{sim} = \theta_{t}$  
  \FOR {$j = 1$ to $T_{sim}$}
  \STATE Make a step with DRS gradient $\theta_{sim} = opt_{sim}.step(\theta_{sim}, \nabla f_{drs}(\theta_{sim}))$
  \ENDFOR
  \STATE Compute a simulation descent direction $g_{surrogate} = \theta_{sim} - \theta_{t}$
  \STATE
  \STATE \textit{// Guided-ES Part}
  \STATE Update low-dimensional guiding subspace $U$ with the $g_{surrogate}$
  \STATE Define search covariance $\Sigma = \frac{\alpha}{n}I + \frac{1-\alpha}{k}UU^{T}$
  \FOR {$i = 1$ to $P$}
   \STATE Sample perturbation $\epsilon_{i} \sim \mathcal{N}(0, \sigma^{2}\Sigma)$
   \STATE Compute antithetic pair of losses $f(\theta_{t} + \epsilon_{i})$ and $f(\theta_{t} - \epsilon_{i})$
  \ENDFOR
 \STATE Compute Guided ES gradient estimate $g = \frac{\beta}{2\sigma^{2}P} \sum_{i=1}^{P}\epsilon_{i}[f(\theta_{t} + \epsilon_{i}) - f(\theta_{t} - \epsilon_{i})]$
 \STATE Update parameters using given optimizer $\theta_{t+1} = opt_{real}.step(\theta_{t}, g)$
 \ENDFOR
 \RETURN $\theta_{T}$ 
 \end{algorithmic} 
\end{algorithm*}

\newpage
\subsection{Environments}

\subsubsection{Pendulum}

\begin{figure}[!htbp]
  \label{fig:pendulum}
  \centering
  \includegraphics[width=0.9\columnwidth]{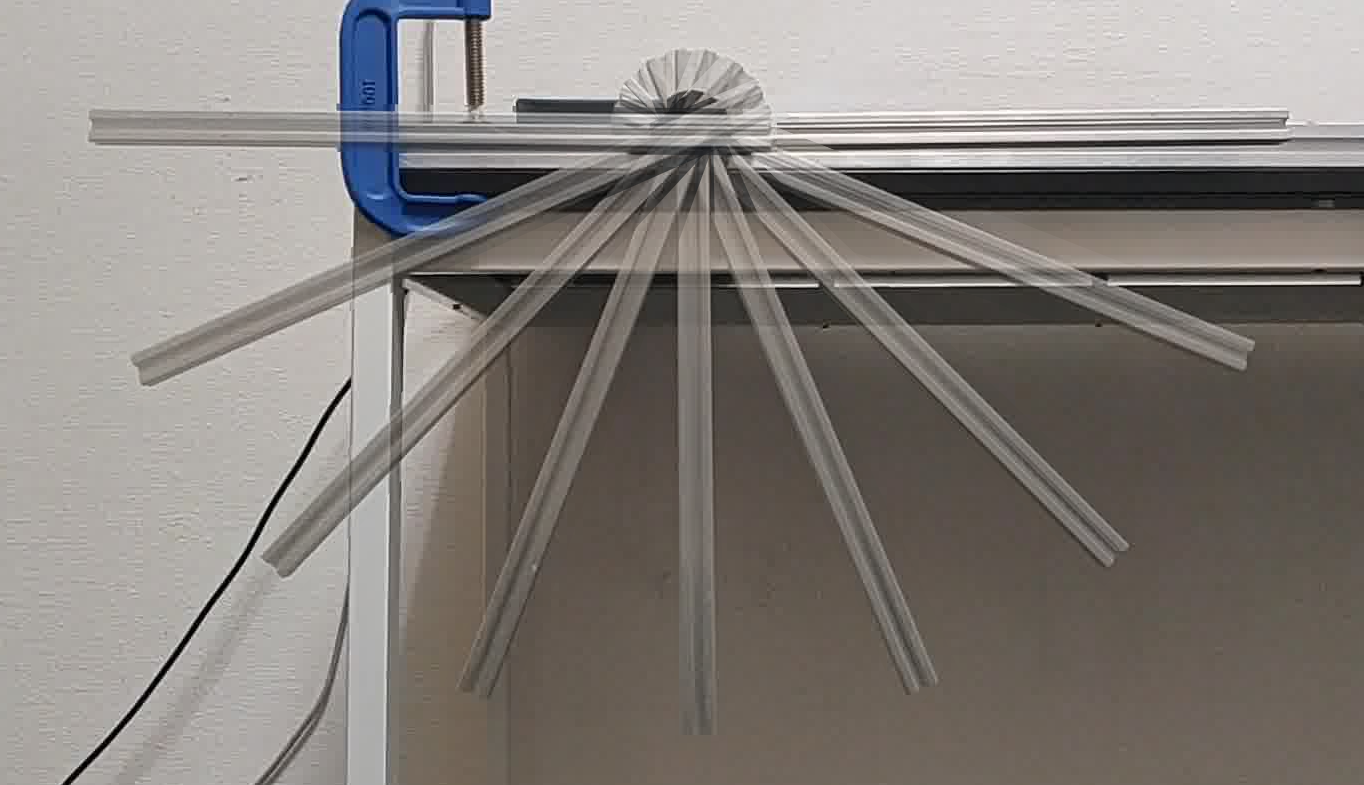}
  \caption{Pendulum robot. The goal is to start swinging at 180 degrees. The robot starts pointing downward at zero velocity and acceleration.}
\end{figure}

\textbf{Cost Function}

\begin{gather*}
 Cost = \sum_{t=0}^{Horizon} (Energy_{t} - TargetEnergy)^{2}\\
 Energy_{t} = 0.5*J*velocity^{2} + mgl*sin(angle)\\
 TargetEnergy = mgl*sin(2 \pi)
\end{gather*}

\subsubsection{Mass-Spring}

\begin{figure}[!h]
\vskip 0.2in
\begin{center}
\centerline{\includegraphics[scale=0.25]{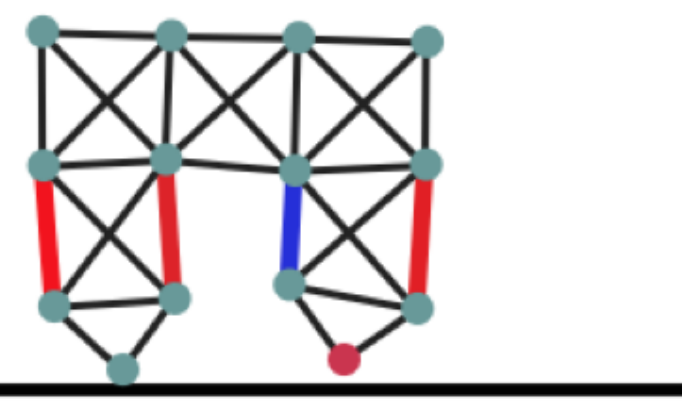}}
\caption{Mass-Spring robot environment. The robot should advance to the right as far as possible in a fixed amount of time. The actuation is done by varying the lengths of the springs colored in red or blue. One simulation corresponds to 8 seconds of real-time. For videos check Differentiable Mass-Spring Simulator section at \url{https://github.com/yuanming-hu/difftaichi}}
\label{fig:mass_spring}
\end{center}
\end{figure}

The environment is depicted in Figure \ref{fig:mass_spring}, the objective function is defined as a difference between the center of mass positions for first and last timesteps projected at $x$.

\end{document}